\documentclass[11pt,a4paper]{article}
\usepackage{times,latexsym}
\usepackage{url}
\usepackage[T1]{fontenc}

    \usepackage[acceptedWithA]{tacl2018v2}
\usepackage[]{tacl2018v2}
\usepackage{graphicx} 
\usepackage{subfigure}
\usepackage{booktabs}
\usepackage{multirow}
\usepackage[namelimits]{amsmath} 
\usepackage{amssymb}           
\usepackage{amsfonts}            
\usepackage{mathrsfs}    
\usepackage{color}

\usepackage{xspace,mfirstuc,tabulary}

\newif\iftaclinstructions
\taclinstructionsfalse 
\iftaclinstructions
\renewcommand{\confidential}{}
\renewcommand{\anonsubtext}{(No author info supplied here, for consistency with TACL-submission anonymization requirements)}
\newcommand{\instr}
\fi

\iftaclpubformat 

\else

\fi

\title{BURT: BERT-inspired Universal Representation from Twin Structure}

\author{
 Yian Li\textsuperscript{\rm 1,2,3},
 Hai Zhao\textsuperscript{\rm 1,2,3,\thanks{Corresponding author. This paper was partially supported by National Key Research and Development Program of China (No. 2017YFB0304100) and Key Projects of National Natural Science Foundation of China (U1836222 and 61733011).}}
 \\
 \textsuperscript{\rm 1}Department of Computer Science and Engineering, Shanghai Jiao Tong University\\
 \textsuperscript{\rm 2}Key Laboratory of Shanghai Education Commission for Intelligent Interaction\\
 and Cognitive Engineering, Shanghai Jiao Tong University, Shanghai, China\\
 \textsuperscript{\rm 3}MoE Key Lab of Artificial Intelligence, AI Institute, Shanghai Jiao Tong University\\
 {\tt liya19@sjtu.edu.cn, zhaohai@cs.sjtu.edu.cn} \\
}

\date{}

\begin{document}
\maketitle
\begin{abstract}
Pre-trained contextualized language models such as BERT \citep{bert} have shown great effectiveness in a wide range of downstream Natural Language Processing (NLP) tasks. However, the effective representations offered by the models target at each token inside a sequence rather than each sequence and the fine-tuning step involves the input of both sequences at one time, leading to unsatisfying representations of various sequences with different granularities. Especially, as sentence-level representations taken as the full training context in these models, there comes inferior performance on lower-level linguistic units (phrases and words). In this work, we present BURT (BERT inspired Universal Representation from Twin Structure) that is capable of generating universal, fixed-size representations for input sequences of any granularity, i.e., words, phrases, and sentences, using a large scale of natural language inference and paraphrase data with multiple training objectives. Our proposed BURT adopts the Siamese network, learning sentence-level representations from natural language inference dataset and word/phrase-level representations from paraphrasing dataset, respectively. We evaluate BURT across different granularities of text similarity tasks, including STS tasks, SemEval2013 Task 5(a) and some commonly used word similarity tasks, where BURT substantially outperforms other representation models on sentence-level datasets and achieves significant improvements in word/phrase-level representation.
\end{abstract}

\section{Introduction}
Representing words, phrases and sentences as low-dimensional dense vectors has always been the key to many Natural Language Processing (NLP) tasks. Previous language representation learning methods can be divided into two different categories based on the language unit they focus on, and therefore are suitable for different situations. High-quality word vectors derived by word embedding models \citep{word2vec,glove,fasttext} are good at measuring syntactic and semantic word similarities and significantly benefit a lot of NLP models. Later proposed sentence encoders \citep{infersent,gensen,use} aim to learn generalized fixed-length sentence representations in supervised or multi-task manners, obtaining substantial results on multiple downstream tasks. Nevertheless, these models focus on either words or sentences, achieving encouraging performance at one level of linguistic unit but less satisfactory results at other levels. 

Recently, contextualized representations with a language model training objective such as ELMo, OpenAI GPT, BERT, XLNet and ALBERT \citep{elmo,gpt-1,bert,xlnet} \citep{albert} are expected to capture complex features (syntax and semantics) for sequences of any length. Especially, BERT improves the pre-training and fine-tuning scenario, obtaining new state-of-the-art results on multiple sentence-level tasks. On the basis of BERT, ALBERT introduces three techniques to reduce memory consumption and training time: decomposing embedding parameters into smaller matrices, sharing parameters cross layers and replacing the next sentence prediction (NSP) training objective with the sentence-order prediction (SOP). In the fine-tuning procedure of both models, the \texttt{[CLS]} token is considered to be the representation of the input sentence pair. Despite its effectiveness, these representations are still token-based and the model requires both sequences to be encoded at one time, leading to unsatisfying representation of an individual sequence. Most importantly, there is a huge gap in representing linguistic units of different granularities. The model ability of handling lower-level linguistic units such as phrases and words is not as good as pre-trained word embeddings.

In this paper, we propose BURT (BERT inspired Universal Representation from Twin Structure) to learn universal representations for different grained linguistic units (including words, phrases and sentences) through multi-task supervised training on two kinds of datasets: NLI \citep{snli,multinli} and the Paraphrase Database (PPDB) \citep{ppdb}. The former is usually used as a sentence-pair classification task to develop semantic representations of sentences. The latter contains a large number of paraphrases, which in our experiments are considered as word/phrase-level paraphrase identification and pairwise text classification tasks. In order to let BURT learn the representation of a single sequence, we adopt the Siamese neural network where each word, phrase and sentence are encoded separately through the twin networks, and then transformed into a fixed-length vector by mean-pooling. Finally, for each sequence pair, the concatenation of the two vectors is fed into a softmax layer for classification. As experiments reveal, our multi-task learning framework combines the characteristics of different training objectives with respect to linguistic units of different granularities.

Our BURT is evaluated on multiple levels of semantic similarity tasks. In addition to standard datasets, we sample pairs of phrases 
from the Paraphrase Database to construct an additional phrase similarity test set. Results show that BURT substantially outperforms sentence representation models including Skip-thought vectors \citep{skip-thought}, InferSent \citep{infersent}, and GenSen \citep{gensen} on seven STS test sets and two phrase-level similarity tasks. Evaluation on word similarity datasets such as SimLex, WS-353, MEN and SCWS also demonstrates that BURT is better at encoding words than other sentence representation models, and even surpasses pre-trained word embedding models by 10.5 points Spearman's correlation on SimLex. 

Generally, BURT can be used as a universal encoder that produces fixed-size representations for different grained input sequences of any granularity without additional training for specific tasks. Moreover, it can be further fine-tuned for downstream tasks by simply adding an decoding layer on top of the model with few task-specific parameters.

\begin{figure*}[t]
\centering
\includegraphics[width=\textwidth]{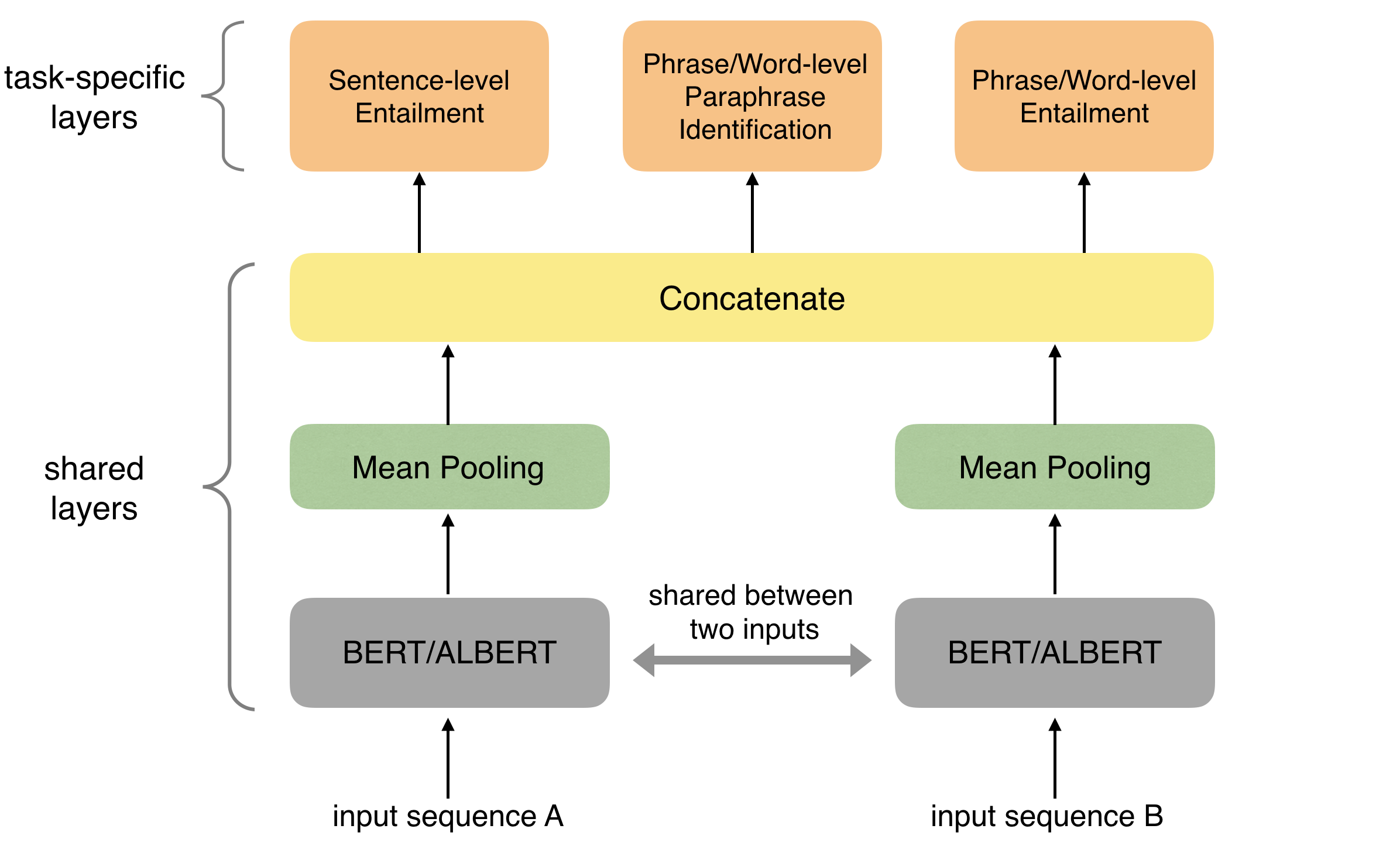}
\caption{Illustration of the architecture of BURT.}
\label{model}
\end{figure*}

\section{Related Work}
Representing words as real-valued dense vectors is a core technology of deep learning in NLP. Trained on massive unlabeled corpus, word embedding models \citep{word2vec,glove,fasttext} map words into a vector space where similar words have similar latent representations. Pre-trained word vectors are well known for their good performance on word similarity tasks, while they are less satisfactory in representing phrases and sentences. Different from the above mentioned static word embedding models, ELMo \citep{elmo} attempts to learn context-dependent word representations through a two-layer bi-directional LSTM network, where each word is allowed to have different representations according to its contexts. The embedding of each token is the concatenation of hidden states from both directions. Nevertheless, most NLP tasks require representations for higher levels of linguistic units such as phrases and sentences.

Generally, phrase embeddings are more difficult to learn than word embeddings for the former has much more diversities in forms and types than the latter. One approach is to treat each phrase as an individual unit and learn its embedding using the same technique for words \citep{phrase2vec}. However, this method requires the preprocess step of extracting frequent phrases in the corpus, and may suffer from data sparsity. Therefore, embeddings learned in this way are not able to truly represent the meaning of phrases. Since distributed representation of words is a powerful technique that has already been used as prior knowledge for all kinds of NLP tasks, one straightforward and simple idea to obtain a phrase representation is to combine the embeddings of all the words in it. To preserve word order and better capture linguistic information, \citet{compose_phrase} come up with complex composition functions rather than simply averaging word embeddings. Based on the analysis of the impact of training data on phrase embeddings, \citet{gru_phrase} propose to train their pairwise-GRU network utilizing large-scaled paraphrase database.

In recent years, more and more researchers have focused on sentence representations since they are widely used in various applications such as information retrieval, sentiment analysis and question answering. The quality of sentence embeddings are usually evaluated in a wide range of downstream NLP tasks. One simple but powerful baseline for learning sentence embeddings is to represent sentence as a weighted sum of word vectors. Inspired by the skip-gram algorithm \citep{word2vec}, the SkipThought model \citep{skip-thought}, where both the encoder and decoder are based on Recurrent Neural Network (RNN), is designed to predict the surrounding sentences for an given passage. \citet{quick-thought} improve the model structure by replacing the decoder with a classifier that distinguishes contexts from other sentences. Rather than unsupervised training, InferSent \citep{infersent} is designed as a bi-directional LSTM sentence encoder that is trained on the Stanford Natural Language Inference (SNLI) dataset. \citet{gensen} introduce a multi-task framework that combines different training objectives and report considerable improvements in downstream tasks even in low-resource settings. To encode a sentence, researchers turn their encoder architecture from RNN as used in SkipThought and InferSent to the Transformer \citep{transformer} which relies completely on attention mechanism to perform Seq2Seq training. \citet{use} develop the Universal Sentence Encoder and explore two variants of model settings: the Transformer architecture and the deep averaging network (DAN) \citep{dan} for encoding sentences. 
 
Most recently, Transformer-based language models such as OpenAI GPT, BERT, Transformer-xl, XLNet and ALBERT \citep{gpt-1,gpt-2,bert,transformer-xl,xlnet} \citep{albert} play an increasingly important role in NLP. Unlike feature-based representation methods, BERT follows the the fine-tuning approach where the model is first trained on a large amount of unlabeled data including two training targets: the masked language model (MLM) and the next sentence prediction (NSP) tasks. Then, the pre-trained model can be easily applied to a wide range of downstream tasks through fine-tuning. During the fine-tuning step, two sentences are concatenated and fed into the input layer and the contextualized embedding of the special token \texttt{[CLS]} added in front of every input is considered as the representation of the sequence pair. \citet{multidnn} further improve the performance on ten downstream tasks by fine-tuning BERT through multiple training objectives. These deep neural models are theoretically capable of representing sequences of arbitrary lengths, while experiments show that they perform unsatisfactorily on word and phrase similarity tasks.

\begin{table*}[t]
  \centering
  \resizebox{\textwidth}{!}{
  \renewcommand{\multirowsetup}{\centering}
  \begin{tabular}{llrrrl}\toprule
  Target&Paraphrase&Google \textit{n}-gram Sim&AGiga Sim&\textit{equivalence} score&Entailment label\\\midrule
  \textit{hundreds}&\textit{thousands}&0.&0.92851&0.000457&\textit{independent}\\\midrule
  \textit{welcomes information that}&\textit{welcomes the fact that}&0.16445&0.64532&0.227435&\textit{independent}\\\hline
  \textit{the results of the work}&\textit{the outcome of the work}&0.51793.&0.95426&0.442545&\textit{entailment}\\\hline
  \textit{and the objectives of the}&\textit{and purpose of the}&0.34082&0.66921&0.286791&\textit{entailment}\\\hline
  \textit{different parts of the world}&\textit{various parts of the world}&0.72741&0.97907&0.520898&\textit{equivalence}\\\hline
  \textit{drawn the attention of the}&\textit{drew the attention of the}&0.72301&0.92588&0.509006&\textit{equivalence}\\\bottomrule
  \end{tabular}}
  \caption{Examples from the PPDB database. Pairs of phrases and words are annotated  with similarities computed from the Google $n$-grams and the Annotated Gigaword corpus, entailment labels and the score for each label.}
   \label{examples}
\end{table*}

\section{Methodology}
The architecture of our universal representation model is shown in Figure \ref{model}. Our BURT follows the Siamese neural network where the twin networks share the same parameters. The lower layers are initialized with either $\rm BERT_{base}$ or $\rm ALBERT_{base}$ which contains 12 Transformer blocks \citep{transformer}, 12 self-attention heads with hidden states of dimension 768. The top layers are task-specific decoders, each consisting of a fully-connected layer followed by a softmax classification layer. We add an additional mean-pooling layer in between to convert a series of hidden states into one fixed-size vector so that the representation of each input is independent of its length. Different from the default fine-tuning procedure of BERT and ALBERT where two sentences are concatenated and encoded as a whole sequence, BURT processes and encodes each word, phrase and sentence separately. The model is trained on two kinds of datasets with regard to different levels of linguistic unit on three tasks. In the following subsections, we present a detailed introduction of the datasets and training objectives.

\subsection{Datasets}
\paragraph{SNLI and Multi-Genre NLI}
The Stanford Natural Language Inference (SNLI) \citep{snli} and the Multi-Genre NLI Corpus \citep{multinli} are sentence-level datasets that are frequently used to improve and evaluate the performance of sentence representation models. The former consists of 570k sentence pairs that are manually annotated with the labels \textit{entailment}, \textit{contradiction}, and \textit{neutral}. The latter is a collection of 433k sentence pairs annotated with textual entailment information. Both datasets are distributed in the same formats except the latter is derived from multiple distinct genres. Therefore, in our experiment, these two corpora are combined and serve as a single dataset for the sentence-level natural language inference task during training. The training and validation sets contain 942k and 29k sentence pairs, respectively.

\paragraph{PPDB}
The Paraphrase Database (PPDB) \citep{ppdb} contains millions of multilingual paraphrases that are automatically extracted from bilingual parallel corpora. Each pair includes one target phrase or word and its paraphrase, companied with entailment information and their similarity score computed from the Google $n$-grams and the Annotated Gigaword corpus. Relationships between pairs fall into six categories: \textit{Equivalence}, \textit{ForwardEntailment}, \textit{ReverseEntailment}, \textit{Independent}, \textit{Exclusion} and \textit{OtherRelated}. Pairs are marked with scores indicating the probabilities that they belong to each of the above six categories. The PPDB database is divided into six sizes, from S up to XXXL, and contains three types of paraphrases: \textit{lexical}, \textit{phrasal}, and \textit{syntactic}. To improve the model ability of representing phrases and words, we take advantage of the \textit{phrasal} dataset with S size, which consists of 1.53 million multiword to single/multiword pairs. We apply a preprocessing step to filter and normalize the data for the phrase and word level tasks. Specifically, pairs tagged with \textit{Exclusion} or \textit{OtherRelated} are removed, and both \textit{ForwardEntailment} and \textit{ReverseEntailment} are treated as \textit{entailment} since our model structure is symmetrical. Finally, we randomly select 354k pairs from each of the three labels: \textit{equivalence}, \textit{entailment} and \textit{independent}, resulting in a total of 1.06 million examples.

\subsection{Training Objectives}
Different from the two training targets of BERT, we introduce three training objectives for BURT as follows to further enhance the output universal representations.

\paragraph{Sentence-level Natural Language Inference}
Natural Language Inference (NLI) is a pairwise classification problem that is to identify the relationship between a premise $P = \{p_1, p_2, \dots, p_{l_P}\}$ and a hypothesis $H = \{h_1, h_2, \dots, h_{l_H}\}$ from \textit{entailment}, \textit{contradiction}, and \textit{neutral}, where $l_P$ and $l_H$ are numbers of tokens in $P$ and $H$, respectively. BURT is trained on the collection of SNLI and Multi-Genre NLI corpora to perform sentence-level encoding. Different from the default preprocess procedure of BERT and ALBERT, the premise and hypothesis are tokenized and encoded separately, resulting in two fixed-length vectors $u$ and $v$. Both sentences share the same set of model parameters during the encoding procedure. We then compute $[u; v; |u-v|]$, which is the concatenation of the premise and hypothesis representations and the absolute value of their difference, and finally feed it to a fully-connected layer followed by a 3-way softmax classification layer. The probability that a sentence pair is labeled as class $a$ is predicted as:
\begin{equation*}
    P(a|P, H) = softmax(W_1^\top \cdot [u; v; |u-v|])
\end{equation*}

\paragraph{Phrase/word-level Paraphrase Identification}
In order to map lower-level linguistic units into the same vector space as sentences, BURT is trained to distinguish between paraphrases and non-paraphrases using a large number of phrase and word pairs from the PPDB dataset. Each paraphrase pair involves a target $T = \{t_1, t_2, \dots, t_{l_T}\}$ and its paraphrase $S_0 = \{s_1, s_2, \dots, s_{l_{S_0}}\}$, where $l_T$ and $l_{S_0}$ are numbers of tokens in $T$ and $S_0$, respectively. Each target or its paraphrase is either a single word or a phrase composed of up to 6 words. Similar to \citet{gru_phrase}, we use the negative sampling strategy \citep{phrase2vec} to reconstruct the dataset. For each target $T$, we randomly sample $k$ sequences $\{S_1, S_2, \dots, S_k\}$ from the dataset and annotate pairs $\{(T, S_1), (T, S_2), \dots, (T, S_k)\}$ with negative labels, indicating they are not paraphrases. The encoding and predicting steps are the same as mentioned in the previous paragraph. The relationship $b$ between $T$ and $S$ is predicted by a logistic regression with softmax: 
\begin{equation*}
    P(b|T, S) = softmax(W_2^\top \cdot [u; v; |u-v|])
\end{equation*}

\paragraph{Phrase/word-level Pairwise Text Classification}
Apart from the paraphrase identification task, we design a word/phrase-level pairwise text classification task to make use of the phrasal entailment information in the PPDB dataset. For each paraphrase pair $(A, B)$, BURT is trained to recognize from three types of relationships: \textit{equivalence}, \textit{entailment} and \textit{independent}. This task is more challenging than the previous one, because the model tries to capture the degree of similarity between phrases and words while words are considered dissimilar even if they are closely related. Examples of paraphrase pairs with different entailment labels and their similarity scores are presented in Table \ref{examples}, where "\textit{hundreds}" and "\textit{thousands}" are labeled as \textit{independent}. A one-layer classifier is used to determine the entailment label $c$ for each pair:
\begin{equation*}
    P(c|A, B) = softmax(W_3^\top \cdot [u; v; |u-v|])
\end{equation*}

\subsection{Training details}
BURT contains one shared encoder with 12 Transformer blocks, 12 self-attention heads and three task-specific decoders. The dimension of hidden states are 768.  In each iteration we train batches from three tasks in turn and make sure that the sentence-level task is trained every 2 batches. We use the Adam optimizer with $\beta_1=0.9$, $\beta_2=0.98$, and $\epsilon=10^{-9}$. We perform warmup over the first 10\% training data and linearly decay learning rate. The batch size is 16 and the dropout rate is 0.1.

\begin{table*}[t]
  \centering
  \renewcommand{\multirowsetup}{\centering}
  \begin{tabular}{l|rrrrrrrr}\toprule
    Method&STS12&STS13&STS14&STS15&STS16&STS B&SICK-R&Avg. \\\midrule\midrule
    
    \multicolumn{9}{l}{\textit{pre-trained word embedding models}}\\
    Avg. GloVe&53.3&50.8&55.6&59.2&57.9&63.0&71.8&58.8\\
    Avg. FastText&58.8&58.8&63.4&69.1&68.2&68.3&73.0&65.7\\\midrule\midrule
    
    \multicolumn{9}{l}{\textit{sentence representation models}}\\
    SkipThought&44.3&30.7&39.1&46.7&54.2&73.7&79.2&52.6\\
    InferSent-2&62.9&56.1&66.4&74.0&72.9&78.5&83.1&70.5\\
    $\rm GenSen_{LAST}$&60.9&55.6&62.8&73.5&66.6&78.6&82.6&68.6\\\midrule\midrule
    
    \multicolumn{9}{l}{\textit{pre-trained contextualized language models}}\\
    $\rm BERT_{CLS}$&32.5&24.0&28.5&35.5&51.1&50.4&64.2&60.6\\
    $\rm BERT_{MAX}$&47.9&45.3&52.6&60.8&60.9&61.3&71.2&57.1\\
    $\rm BERT_{MEAN}$&50.1&52.9&54.9&63.4&64.9&64.5&73.5&40.9\\\midrule\midrule
    
    \multicolumn{9}{l}{\textit{our methods}}\\
    ALBURT &67.2&72.6&71.4&77.8&73.5&79.7&83.5&75.1 \\
    BURT &\textbf{69.9}&\textbf{73.7}&\textbf{72.8}&\textbf{78.5}&\textbf{73.6}&\textbf{84.3}&\textbf{84.8}&\textbf{76.8}\\\bottomrule
  \end{tabular}
  \caption{Performance of BURT/ALBURT and baseline models on sentence similarity tasks. BURT and ALBURT are initialized with $\rm BERT_{base}$ and $\rm ALBERT_{base}$, respectively. "SemEval" stands for the SemEval Task 5(a). The subscript refers to the pooling strategy used to obtain fixed-length representations. The best results are in bold.}
   \label{evaluation}
\end{table*}

\begin{table*}[t]
  \centering
  \renewcommand{\multirowsetup}{\centering}
  \begin{tabular}{l|rr|rrrrrr}\toprule
    \multirow{2}{1in}{Method}&\multicolumn{2}{c}{Phrases}&\multicolumn{6}{c}{Words}\\
    &PPDB& SemEval&SimLex&WS-sim&WS-rel&MEN&SCWS&Avg.\\\midrule\midrule
    
    \multicolumn{9}{l}{\textit{pre-trained word embedding models}}\\
    Avg. GloVe&35.9&68.2&40.8&80.2&64.4&80.5&62.9&65.8\\
    Avg. FastText&32.2&68.9&50.3&\textbf{83.4}&\textbf{73.4}&\textbf{84.6}&\textbf{69.4}&\textbf{72.2}\\\midrule\midrule
    
    \multicolumn{9}{l}{\textit{sentence representation models}}\\
    SkipThought&37.1&66.4&35.1&61.3&42.2&57.9&58.4&51.0\\
    InferSent-2&36.3&78.2&55.9&71.1&44.4&77.4&61.4&62.0\\
    $\rm GenSen_{LAST}$&48.6&64.3&50.0&56.0&34.0&59.3&59.0&51.7\\\midrule\midrule
    
    \multicolumn{9}{l}{\textit{pre-trained contextualized language models}}\\
    $\rm BERT_{CLS}$&30.2&75.9&7.3&23.1&1.8&19.1&28.1&14.4\\
    $\rm BERT_{MAX}$&44.9&77.0&16.7&30.7&14.6&26.8&34.5&24.6\\
    $\rm BERT_{MEAN}$&42.8&71.2&13.1&3.0&11.2&21.8&23.2&15.9\\\midrule\midrule
    
    \multicolumn{9}{l}{\textit{our methods}}\\
    ALBURT &66.7&75.9&59.6&\underline{73.9}&\underline{61.0}&69.5&\underline{63.9}&\underline{65.6} \\
    BURT &\textbf{70.6}&\textbf{79.3}&\textbf{60.8}&\underline{71.5}&\underline{55.7}&68.5&\underline{62.3}&\underline{63.7}\\\bottomrule
  \end{tabular}
  \caption{Performance of BURT/ALBURT and baseline models on phrase and word similarity tasks. "WS-rel" = subtask of WS353 where words are related. "WS-sim" = subtask of WS353 where words are similar. The best results are in bold. Underlined cells shows tasks where BURT outperforms all the sentence representation models and the pre-trained BERT.}
   \label{evaluation_2}
\end{table*}

\section{Evaluation}
BURT is evaluated on several text similarity tasks with respect to different levels of linguistic units, i.e, sentences, phrases and words. We use cosine similarity to measure the distance between two sequences:
\begin{equation*}
    cosine(u, v) = \frac{u\cdot v}{\|u\|\|v\|}
\end{equation*}
where $u\cdot v$ is the dot product of $u$ and $v$, and $\|u\|$ is the $\ell_2$-norm of the vector. Then Spearman's correlation between these cosine similarities and golden labels is computed to investigate  how much semantic information is captured by our model.

\paragraph{Baselines}
We use several currently popular word embedding models and sentence representation models as our baselines. Pre-trained word embeddings used in this work include GloVe \citep{glove} and FastText \citep{fasttext}, both of which have dimensionality 300. They represent a phrase or a sentence by averaging all the vectors of words it contains. SkipThought \citep{skip-thought} is an encoder-decoder structure trained in an unsupervised way, where both the encoder and decoder are composed of GRU units. Two versions of trained models are provided: unidirectional and bidirectional. In our experiment, the concatenation of the last hidden states produced by the two models are considered to be the representation of the input sequence. The dimensionality is 4800. InferSent \citep{infersent} is a bidirectional LSTM encoder trained on the SNLI dataset with a max-pooling layer. It has two versions: InferSent1 with GloVe vectors and InferSent2 with FastText vectors. The latter is evaluated in our experiment. The output vector is 4096-dimensional. GenSen \citep{gensen} is trained through a multi-task learning framework to learn general-purpose representations for sentences. The encoder is a bidirectional GRU. Multiple trained models of different training settings are publicly available and we choose the single-layer models that are trained on skip-thought vectors, neural machine translation, constituency parsing and natural language inference tasks. Since the last hidden states work better than max-pooling on the STS datasets, they are used as representations for sequences in our experiment. The dimensionality of the embeddings is 4096. In addition, we examine the pre-trained model $\rm BERT_{base}$ using different pooling strategies: mean-pooling, max-pooling and the \texttt{[CLS]} token. The dimension is 768.

\subsection{Sentence-level evaluation}
We evaluate the model performance of encoding sentences on SentEval \citep{senteval}, including datasets that require training: the STS benchmark \citep{stsb} and the SICK-Relatedness dataset \citep{sick}, and datasets that do not require training: the STS tasks 2012-2016 \citep{sts2012,sts2013,sts2014,sts2015,sts2016}. For BURT and the baseline models, we encode each sentence into a fixed-length vector with their corresponding encoders and pooling layers, and then compute cosine similarity between each sentence pair. For the STS benchmark and the SICK-Relatedness dataset, each model is first trained on a regression task and then evaluated on the test set. The results are displayed in Table \ref{evaluation}.

According to the results, BURT outperforms all the baseline models on seven evaluated datasets, achieving significant improvements on sentence similarity tasks. As expected, averaging pre-trained word vectors leads to inferior sentence embeddings than well-trained sentence representation models. 
Among the three RNN-based sentence encoders, InferSent and GenSen obtain higher correlation than SkipThought, but not as good as our model. The last five rows in the table indicate that our multi-task framework essentially improves the quality of sentence embeddings. With an average score of 76.8, BURT outperforms ALBURT of the same size. Benefiting from the training of SNLI and Multi-Genre NLI corpora using the Siamese structure, BURT can be directly used as a sentence encoder to extract features for downstream tasks without further training of the parameters. Besides, it can also be efficiently fine-tuned to generate sentence vectors for specific tasks by adding an additional decoding layer such as a softmax classifier.

\begin{table*}[t]
  \centering
  \resizebox{\textwidth}{!}{
  \renewcommand{\multirowsetup}{\centering}
  \begin{tabular}{l|rrr|rr|rrrrrr}\toprule
    \multirow{2}{1in}{}&\multicolumn{3}{c}{Setences}&\multicolumn{2}{c}{Phrases}&\multicolumn{6}{c}{Words}\\
    &STS* &STS B&SICK-R&PPDB& SemEval&SimLex&WS-sim&WS-rel&MEN&SCWS&Avg.\\\midrule\midrule
    
    \multicolumn{12}{l}{\textit{Traning Objectives}}\\
    NLI&\large 73.5&\large \textbf{85.4}&\large \textbf{85.2}&\large  31.5&\large 77.3&\large 57.9&\large 24.9&\large 30.0&\large 55.1&\large 47.6&\large 30.8\\
    NLI+PI&\large 73.0&\large 82.7&\large 84.7&\large \large 40.4&\large 77.3&\large 59.6&\large 69.4&\large \textbf{57.4}&\large 67.0&\large \textbf{64.4}&\large 63.6\\
    NLI+PTC&\large 72.4&\large 84.2&\large 84.9&\large \textbf{76.9}&\large 78.3&\large 41.2&\large 56.1&\large 30.0&\large 48.4&\large 55.4&\large 46.2\\
    NLI+PI+PTC&\large \textbf{73.7}&\large 84.3&\large 84.8&\large 70.6&\large \textbf{79.3}&\large \textbf{60.8}&\large \textbf{71.5}&\large 55.7&\large \textbf{68.5}&\large 62.3&\large \textbf{63.7}\\\midrule
    
    \multicolumn{12}{l}{\textit{Pooling Strategies}}\\
    CLS&\large 71.2&\large 82.6&\large \textbf{85.0}&\large 68.1&\large 78.1&\large 48.0&\large 63.5&\large 42.8&\large 62.7&\large 57.3&\large 54.9\\
    MAX&\large 72.3&\large 78.5&\large 83.6&\large 68.5&\large 77.6&\large 52.4&\large 64.6&\large 46.0&\large 62.0&\large 61.0&\large 57.2\\
    MEAN&\large \textbf{73.7}&\large \textbf{84.3}&\large 84.8&\large \textbf{70.6}&\large \textbf{79.3}&\large \textbf{60.8}&\large \textbf{71.5}&\large \textbf{55.7}&\large \textbf{68.5}&\large \textbf{62.3}&\large \textbf{63.7}\\\midrule
    
    \multicolumn{12}{l}{\textit{Concatenation Methods}}\\
    $[u; v]$&\large 38.2&\large 77.6&\large 81.0&\large 31.3&\large 69.2&\large 7.8&\large 12.5&\large 8.9&\large 19.6&\large 22.8&\large 14.3\\
    $[|u-v|]$&\large 56.6&\large 84.2&\large 84.1&\large 41.5&\large 76.9&\large 50.8&\large 51.6&\large 22.4&\large 49.0&\large 53.9&\large 45.5\\
    $[u*v]$&\large 67.1&\large 81.7&\large 84.2&\large 70.2&\large 75.2&\large 51.4&\large 62.8&\large 42.8&\large 62.9&\large 58.3&\large 55.6\\
    $[u; v; |u-v|]$&\large \textbf{73.7}&\large \textbf{84.3}&\large 84.8&\large \textbf{70.6}&\large \textbf{79.3}&\large \textbf{60.8}&\large \textbf{71.5}&\large \textbf{55.7}&\large \textbf{68.5}&\large \textbf{62.3}&\large \textbf{63.7}\\
    $[u; v; u*v]$&\large 70.2&\large 83.5&\large 84.6&\large 62.3&\large 77.4&\large 53.0&\large 69.0&\large 52.4&\large 64.3&\large 59.3&\large 59.6\\
    $[|u-v|; u*v]$&\large 70.7&\large 83.4&\large 84.8&\large 69.7&\large 77.3&\large 50.3&\large 71.5&\large 46.3&\large 61.0&\large 58.2&\large 57.5\\
    $[u; v; |u-v|; u*v]$&\large 71.9&\large 84.0&\large \textbf{85.0}&\large 62.3&\large 78.0&\large 52.1&\large 64.7&\large 42.9&\large 63.5&\large 58.6&\large 56.4\\\midrule
    
    \multicolumn{12}{l}{\textit{Negative Sampling}}\\
    $k=1$&\large 73.3&\large 84.1&\large \textbf{85.2}&\large 60.6&\large 78.2&\large 60.4&\large 54.6&\large 45.7&\large 61.3&\large 58.9&\large 56.2\\
    $k=3$&\large \textbf{73.7}&\large \textbf{84.3}&\large 84.8&\large 70.6&\large \textbf{79.3}&\large \textbf{60.8}&\large \textbf{71.5}&\large \textbf{55.7}&\large \textbf{68.5}&\large \textbf{62.3}&\large \textbf{63.7}\\
    $k=5$&\large 72.5&\large 83.1&\large 84.9&\large \textbf{72.3}&\large 77.3&\large 54.6&\large 64.6&\large 48.4&\large 63.5&\large 59.8&\large 58.2\\
    $k=7$&\large 71.9&\large 83.1&\large 84.5&\large 69.9&\large 77.9&\large 59.3&\large 68.8&\large 50.9&\large 66.7&\large 60.6&\large 61.3\\\bottomrule
  \end{tabular}
  }
  \caption{Ablation study on training objectives, pooling strategies, concatenation methods and the value of k in negative sampling. Lower layers are initialized with $\rm BERT_{base}$. "STS*" means the average score of STS12 $\sim$ STS16. "NLI", "PI" and "PTC" stands for sentence-level natural language inference, phrase/word-level paraphrase identification and phrase/word-level pairwise text classification tasks, respectively. "PPDB" represents the phrase-level similarity test set extracted from the PPDB database.}
  \label{ablation}
\end{table*}

\subsection{Phrase-level evaluation}
It is not enough to generate high-quality sentence embeddings, BURT is expected to encode semantic information for lower-level linguistic units as well. In this experiment, we perform phrase-level evaluation on SemEval2013 Task 5(a) \citep{semeval2013} which is a task to classify whether a pair of sequences are semantically similar or not. Each pair contains a word and a phrase consisting of multiple words, coming with either a negative or a positive label. Examples like (\textit{megalomania}, \textit{great madness}) are labeled dissimilar although the two sequences are related in certain aspects. Our BURT and baseline models are first fine-tuned on the training set of size 11,722 and then evaluated on 7,814 examples. Parameters of the pre-trained word embeddings, SkipThought, InferSent and GenSen are held fixed during training. Due to limited resource of phrasal semantic datasets, we design a phrase-level semantic similarity test set in the same format as the STS datasets. Specifically, we select pairs from the test set of PPDB and filter out ones containing only single words, resulting in 32,202 phase pairs. In the following experiments, the score of \textit{equivalence} annotated in the original dataset is considered as the relative similarity between the two phrases.

As shown in the first two columns of Table \ref{evaluation_2}, BURT outperforms all the baseline models on SemEval2013 Task 5(a) and PPDB, suggesting that BURT learns semantically coherent phrase embeddings. Note the inconsistent behavior among models on different datasets since the two datasets are distributed differently. Sequences in SemEval2013 Task 5(a) are either single words or two-word phrases. Therefore, GloVe and FastText outperform SkipThought and GenSen on that dataset. Training BURT on $\rm BERT_{base}$ results in higher accuracy than unsupervised models like GloVe, FastText and SkipThought, and is comparable to InferSent and the general-purpose sentence encoder. By first training using a large amount of phrase-level paraphrase data, then fine-tuning on SemEval training data, BURT yields the highest accuracy. On PPDB where most phrases consist more than two words, RNNs and Transformers are more preferable than simply averaging word embeddings. Especially, BURT obtains the highest correlation of 70.6. 

\subsection{Word-level evaluation}
Word-level evaluation are conducted on several commonly used word similarity task datasets: SimLex \citep{simlex999}, WS-353 \citep{ws353}, MEN \citep{men}, SCWS \citep{scws}. As mentioned in \citet{problems}, word similarity is often confused with relatedness in some datasets due to the subjectivity of human annotations. To alleviate this problem, WS-353 is later divided into two sub-datasets \citep{353sim_353rel}: pairs of words that are similar, and pairs of words that are related. For example, "\textit{food}" and "\textit{fruit}" are similar while "\textit{computer}" and "\textit{keyboard}" are related. The newly constructed dataset SimLex aim to explicitly quantifies similarity rather than relatedness, where similar words are annotated with higher scores and related words are considered dissimilar with lower scores. The numbers of word pairs in SimLex, WS-353, MEN and SCWS are 999, 353, 3000 and 2003, respectively. 
Cosine similarity between word vectors are computed and Spearman's correlation is used for evaluation. Results are in the last five columns of Table \ref{evaluation_2}. 

Although pre-trained word embeddings are considered excellent at representing words, BURT obtains the highest correlations on SimLex, 4.9 higher than the best results reported by all the baseline models. Compared with all the three sentence representation models and the pre-trained BERT, BURT yields the best results on 4 out of 5 datasets. The only dataset where InferSent performs better than BURT is MEN. Because InferSent is trained by initializing its embedding layers with pre-trained word vectors, it inherited high-quality word embeddings from FastText to some extent. However, our model is initialized with pre-trained models and fine-tuned using NLI and PPDB datasets, leading to significant improvements in representing words, with an average correlation of 63.7 using BURT and 65.6 using ALBURT compared to only 24.6 using the pre-trained $\rm BERT_{base}$.

\section{Ablation Study}
Our proposed BURT is trained on the NLI and PPDB datasets using three training objectives in terms of different levels of linguistic units, leading to powerful performance for mapping words, phrases and sentences into the same vector space. In this section, we look into how variants of training objectives, pooling and concatenation strategies and some hyperparameters affect model performance and figure out the overall contribution of each module. The lower layers are initialized with $\rm BERT_{base}$ in the following experiments.

\subsection{Training Objective}
We train BURT on different combinations of training objectives to investigate in what aspect does the model benefit from each of them. According to the first block in Table \ref{ablation}, training on the NLI dataset through sentence-level task can effectively improve the quality of sentence embeddings, but it is not helpful enough when it comes to phrases and words. The phrase and word level tasks using the PPDB dataset are able to address this limitation. Especially, the word/phrase-level pairwise text classification task has an positive impact on phrase embeddings. Furthermore, when trained on the paraphrase identification task, BURT is able to produce word embeddings that are almost as good as pre-trained word vectors. By combining the characteristics of different training objectives, our model have an advantage in encoding sentences, meanwhile achieving considerable improvement in phrase-level and word-level tasks.

\subsection{Pooling and Concatenation}
In this subsection, we apply different pooling strategies, including mean-pooling, max-pooling and the \texttt{[CLS]} token, and different feature concatenation methods to BURT. When investigating the former, we use $[u; v; |u-v|]$ as the input feature for classification, and as for the latter, we choose mean-pooling as the default strategy. Results are shown in the second and third blocks in Table \ref{ablation}. In accordance with \citet{sbert}, we find averaging BURT token embeddings outperforms other pooling methods. Besides, Hadamard product $u*v$ is not helpful in our experiment. Generally, BURT is more sensitive to concatenation methods while pooling strategies have a minor influence. For a comprehensive consideration, mean-pooling is preferred, and the concatenation of two vectors along with their absolute difference is more suitable in our experiments than other combinations.

\subsection{Negative Sampling}
In the paraphrase identification task, we randomly select $k$ negative samples for each pair to force the model to identify paraphrases from non-paraphrases. Evidence has shown that the value of $k$ in negative sampling has an impact on phrase and word embeddings \citep{phrase2vec}. When training word embeddings using negative sampling, setting $k$ in the range of 5-20 is recommended for small training datasets, while for large datasets the $k$ can be as small as 2-5. In this ablation experiment, we explore the optimal value of $k$ for our paraphrase identification task. Since the PPDB dataset is extremely large, with more than one million positive pairs, we perform four experiments, in which we only change the value of $k$ and other model settings are maintained the same. The last block in Table \ref{ablation} illustrates results using values of 1, 3, 5 and 7, from which we can conclude that $k=3$ is the best choice. Keeping increasing the value of $k$ has no positive effect on phrase and word embeddings and even decreases the performance in sentence tasks.

\section{Conclusion}
In this work, we propose to learn a universal encoder that maps sequences of different linguistic granularities into the same vector space where similar sequences have similar representations. For this purpose, we present BURT (BERT inspired Universal Representation from Twin Structure), which is trained with three objectives on the NLI and PPDB datasets through a Siamese network. BURT is evaluated on a wide range of similarity tasks with regard to multiple levels of linguistic units (sentences, phrases and words). Overall, our proposed BURT outperforms all the baseline models on sentence and phrase level evaluations, and generates high-quality word vectors that are almost as good as pre-trained word embeddings.

\bibliography{tacl2018}
\bibliographystyle{acl_natbib}

\end{document}